\documentclass[letterpaper, 10 pt, conference]{ieeeconf}  

\IEEEoverridecommandlockouts                              

\usepackage{graphicx}
\usepackage[table,xcdraw]{xcolor}
\usepackage{hyperref}
\usepackage{cite}
\usepackage{amsmath,amssymb,amsfonts}
\usepackage{graphicx}
\usepackage{caption}
\usepackage{textcomp}
\usepackage{xcolor}
\usepackage{booktabs}
\usepackage{graphicx}
\usepackage{url}
\usepackage{subfigure}
\usepackage{algpseudocode}
\usepackage{amsmath}
\usepackage{algorithm2e}
\usepackage{mwe}
\usepackage{balance}
\usepackage{multirow}
\usepackage{diagbox}

\def\BibTeX{{\rm B\kern-.05em{\sc i\kern-.025em b}\kern-.08em
    T\kern-.1667em\lower.7ex\hbox{E}\kern-.125emX}}

\newcommand{\eg}{{\em e.g.,~}}
\newcommand{\ie}{{\em i.e.,~}}

\title{\LARGE \bf 
Kinodynamic Task and Motion Planning using \\ VLM-guided and Interleaved Sampling
}

\author{Minseo Kwon and Young J. Kim
\thanks{The authors are with the Department of Computer Science and Engineering at Ewha Womans University in South Korea  ${\it \{minseo.kwon|kimy\}@ewha.ac.kr}$.}
}

\begin{document}

\maketitle
\thispagestyle{empty}
\pagestyle{empty}

\begin{abstract}
Task and Motion Planning (TAMP) integrates high-level task planning with low-level motion feasibility, but existing methods are costly in long-horizon problems due to excessive motion sampling. While LLMs provide commonsense priors, they lack 3D spatial reasoning and cannot ensure geometric or dynamic feasibility. We propose a kinodynamic TAMP planner based on a {\it hybrid state tree} that uniformly represents symbolic and numeric states during planning, enabling task and motion decisions to be jointly decided. Kinodynamic constraints embedded in the TAMP problem are verified by an off-the-shelf motion planner and physics simulator, and a VLM guides exploring a TAMP solution and backtracks the search based on visual rendering of the states. 
Experiments on the simulated domains and in the real world show $32.14\% \sim 1166.67\%$ increased average success rates compared to traditional and LLM-based TAMP planners and reduced planning time on complex problems, with ablations further highlighting the benefits of VLM backtracking.
More details are available at \href{https://graphics.ewha.ac.kr/kinodynamicTAMP/}{https://graphics.ewha.ac.kr/kinodynamicTAMP/}.
\end{abstract}

\section{INTRODUCTION}
Robotic manipulation tasks, such as tabletop manipulations, require reasoning over both symbolic task decisions and continuous geometric feasibility. A robot must decide which action to perform—such as picking or placing— and which object to grasp, which constitutes a discrete search process. Simultaneously, it must determine grasp poses, feasible end-effector configurations, and collision-free motion trajectories governed by continuous constraints. This class of problems is studied under the framework of Task and Motion Planning (\ie TAMP), which combines high-level task planning with continuous action parameter binding and low-level motion planning \cite{garrett2021integrated}.

When every high-level task plan has a feasible low-level motion solution, {\ie {\it the downward refinement property}}, task plans can be computed independently of geometric constraints and later refined with continuous action parameters. Unfortunately, this assumption rarely holds in practice, where geometric infeasibility can invalidate a task plan. Hence, most TAMP systems interleave discrete and continuous reasoning. 
Early approaches follow two main paradigms: \textit{sequencing-first}, which generates symbolic task plans first and then attempts to satisfy continuous constraints, or \textit{satisfaction-first}, which samples feasible motions first and integrates them into task planning \cite{garrett2021integrated}.
The former suffers from repeated expensive constraint solving when samples are infeasible, while the latter wastes computation on many unused samples \cite{vu2024coast}.

With the emergence of large language models (LLMs), a new class of TAMP solvers has been proposed. These methods leverage the commonsense knowledge and high-level reasoning abilities of pretrained models to generate problem formulations for TAMP \cite{chen2024autotamp}, plausible task sequences \cite{lee2024prime}, and even continuous parameters, such as placement positions and motion failure reasoning \cite{wang2024llmˆ}. 
However, due to the lack of their 3D spatial understanding, LLMs often struggle with precise motion reasoning and cannot reliably validate whether a plan is geometrically or physically feasible. Supplying them with high-dimensional numeric values like 6D poses or motion trajectories yields little benefit \cite{lee2024prime, sharma2023exploring}. 
Furthermore, existing TAMP approaches often ignore inertial or dynamic constraints in the planning environment. This motivates the need to incorporate kinodynamic constraints into our framework, ensuring that generated motions are executable in the real world.

To overcome these limitations, we propose a novel kinodynamic TAMP framework that interleaves a sequencing-first and a satisfaction-first approach, coupling task-level decisions and motion validations at every search-expansion step. This is abstracted through a {\it hybrid search tree} representation, where states and actions are grounded with sampled action parameters and immediately validated in physics simulation. 
An off-the-shelf motion planner and a physics simulator ensure kinodynamic feasibility, such as \eg collisions, kinematics, grasp stability, and object stability. The VLM further evaluates resulting world states to bias exploration toward promising branches and enable backtracking to recover from unexpandable states.
We validate our approach on two simulated and one real-world manipulation domains with varying problem complexity. 
Our method outperforms a traditional domain-independent TAMP planner \cite{garrett2020pddlstream} and an LLM-based TAMP planner \cite{wang2024llmˆ} in terms of success rates and achieves shorter planning times on complex problems, whereas the baselines mostly fail.
We also perform an ablation study to show that VLM guidance significantly improves the quality of backtracking decisions.

Our contributions are summarized as follows:
\begin{itemize}
    \item {We introduce a \textit{hybrid state tree} that unifies symbolic task decisions and instantiating continuous action, \ie a novel \textit{interleaved} formulation of task and motion planning.}
    \item {To realize this abstraction, we combine a top-$k$ symbolic planner for task diversity with a motion planner and a physics simulator, allowing us to check kinodynamic planning constraints.}
    \item We demonstrate that VLMs can be effectively used not only as a forward search heuristic but also for backtracking guidance, improving recovery from failures.
    \item {We evaluate our framework on two simulated manipulation domains, achieving $32.14\% \sim  105.56 \%$ increased average success rates in the Blocksworld domain and $280.00\% \sim  1166.67 \%$ increased average success rates in the Kitchen domain compared to traditional TAMP and LLM-based planners, and provide ablation studies highlighting the impact of VLM guidance.}
    \item We show that our TAMP planner works for a physical robot with kinodynamic constraints, close to the simulated results. 
\end{itemize}


\begin{figure*}
    \centering
    \includegraphics[width=\textwidth]{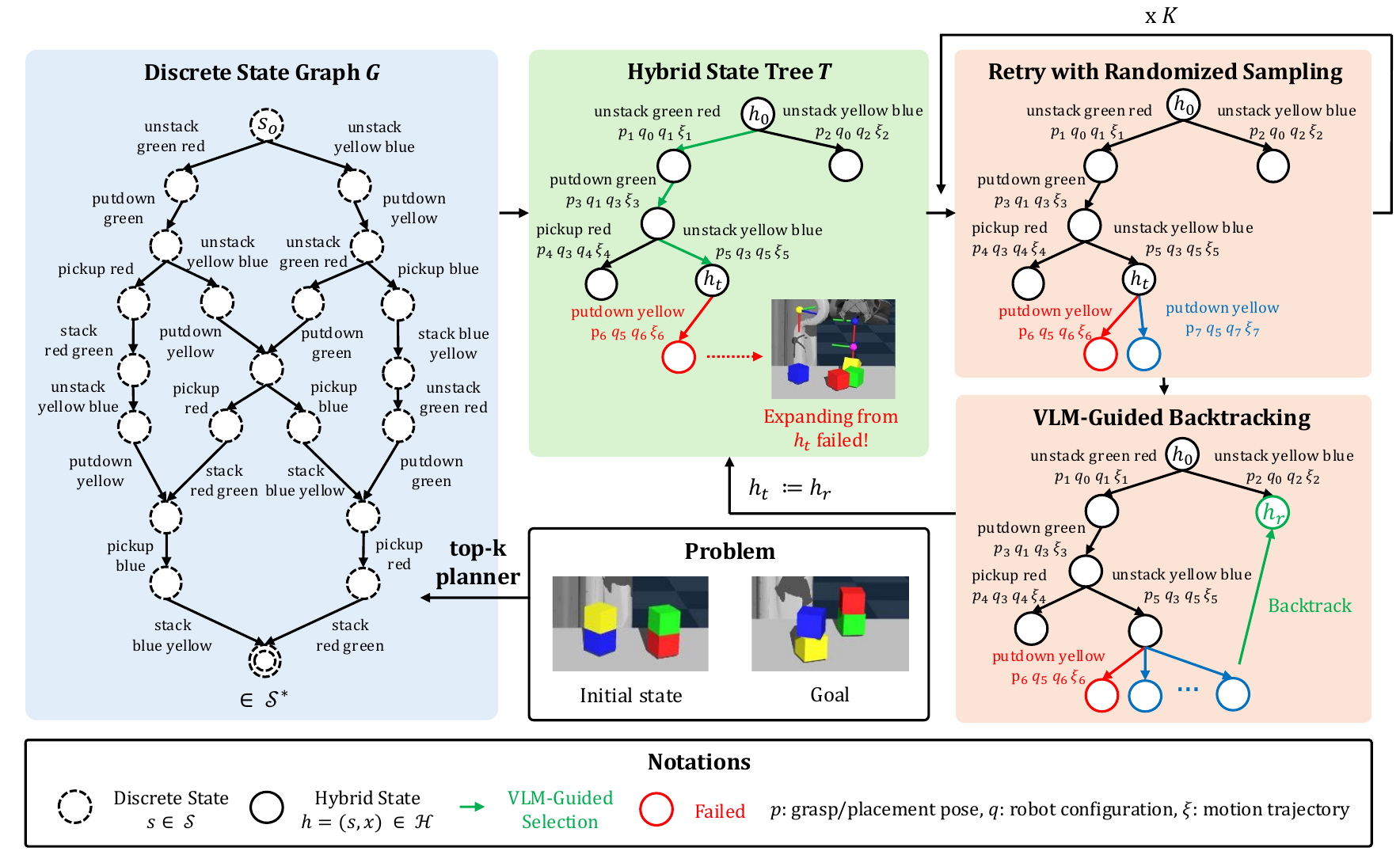}
    \caption{An overview of our approach. Given a problem PDDL and a domain PDDL, a top-$k$ symbolic planner generates a discrete state graph $G$. Guided by $G$, we expand a hybrid state tree $T$ where each edge is expanded by motion planning and validated by physics simulation. If a node $h_t$ fails to expand, we retry random sampling up to $K$ times; if still unsuccessful, we prompt the VLM to predict a backtrack node $h_r$, from which expansion resumes. This process repeats until a goal is found or the timeout is reached.}
    \label{fig:method}
    \vspace{-1.5em}
\end{figure*}

\section{{RELATED WORKS}} \label{sec:relatedwork}

\subsection{Task and Motion Planning (TAMP)}
Traditional TAMP approaches are commonly categorized as sequencing-first, satisfaction-first, or interleaved. 
Sequencing-first methods first compute a symbolic task plan and then attempt to satisfy continuous constraints. Repeatedly solving constraint satisfaction problems after motion failures can be costly: these methods may suffer from planning-time explosion for long horizon tasks \cite{garrett2020pddlstream}, require workspace discretization that limits scalability \cite{dantam2018incremental}, or rely on manually specified constraints that are difficult to generalize across different problems and domains \cite{vu2024coast}. 
Satisfaction-first methods \cite{garrett2018sampling, garrett2018ffrob} generate samples before symbolic search, effective when sampling is cheap. That said, they suffer from combinatorial sample explosion, often producing many unnecessary samples.

Interleaved methods integrate sequencing and action parameter sampling within a unified search. 
For instance, sequencing-first planning can be combined with interleaved reasoning by pruning geometric samples that make the task plan infeasible \cite{lagriffoul2014efficiently}. However, this is limited to refining a single fixed plan skeleton and does not explore diverse task plans. 
More recent frameworks, such as \cite{ren2024extended}, treat each plan skeleton as a top-level branch and employ MCTS for motion binding, but remain limited by random rollout policies that scale poorly in large domains.
Our approach generates a discrete state graph to explore diverse task plans and expands the hybrid tree only with actions that satisfy all constraints, validated by a physics simulator for kinodynamic feasibility.

Learning-based approaches learn symbolic operators \cite{silver2021learning}, or predicates \cite{silver2023predicate} for TAMP. Recent works apply transformers \cite{yang2022sequence} to predict the feasibility of plans or GNNs \cite{khodeir2023learning} to predict the relevance of sampled objects. However, data collection is costly, and generalization is limited, whereas we leverage pre-trained VLMs with commonsense knowledge.

\subsection{LLMs/VLMs for Task and Motion Planning}
LLMs and VLMs have recently been applied to robotic planning, leveraging commonsense reasoning to sequence high-level action primitives \cite{singh2023progprompt, ahn2022can, huang2022inner} or generate robot executable code that captures spatial reasoning and feedback loops \cite{liang2023code}. 
They have also been combined with symbolic planning languages \cite{silver2024generalized, silver2022pddl, liu2023llm+, guan2023leveraging}.
However, these methods are restricted to task-level reasoning, without handling complex geometric or physical constraints.

LLMs and VLMs have been extended beyond simple task planning or skill sequencing to address TAMP problems. 
\cite{ding2023task} employs LLMs to infer spatial relationships between objects, which are then passed to a TAMP planner. 
Other works generate problem formulations with LLMs/VLMs that are solved by external TAMP planners, with corrective re-prompting applied when failures occur \cite{chen2024autotamp, yang2025guiding, siburian2025grounded, kwon2025neuro}. 
Similarly, \cite{guo2025castl} translates natural language constraints into Python APIs for an external TAMP planner, while \cite{kumar2024open} uses VLMs to infer open-world constraints—including both high-level action orders and continuous feasibility checks—through executable code integrated with an external planner.
However, these approaches typically remain limited to coupling LLMs/VLMs with traditional TAMP planners.

\cite{curtis2024trust} frames the search for valid LLM-generated Python programs and their continuous parameters as a constraint satisfaction problem. 
When violations occur, it enforces geometric and physical constraints through iterative re-prompting.
LLMs can also directly produce task and motion plans: \cite{lee2024prime} generates batches of task plans refined with continuous parameters and expanded via MCTS, and \cite{wang2024llmˆ} jointly generates symbolic sequences and continuous parameters, with re-prompting guided by motion planning feedback.
However, unlike our approach, they rely solely on textual feedback for re-prompting and do not use visual cues to evaluate complex kinodynamic constraints during planning and replanning.

\subsection{Kinodynamic Planning}  
Kinodynamic motion planning was first introduced by \cite{donald1993kinodynamic} to address both kinematic constraints, such as obstacle avoidance, and dynamic constraints, including limits on velocity, acceleration, force, and torque, to find time-optimal motion paths. \cite{lavalle2001randomized} later proposed a randomized approach that extended RRTs to kinodynamic settings.
Kinodynamic approaches have also been applied in TAMP. \cite{plaku2010sampling} integrated a sampling-based motion planner with a symbolic planner to jointly solve high-level task reasoning and low-level kinodynamic motion, thereby generating dynamically feasible trajectories. \cite{toussaint2018differentiable} employed differentiable physics engines to formulate planning as a gradient-based optimization problem, enabling reasoning over tool-use domains such as hitting and throwing.  
While classical kinodynamic planners focus on explicitly generating kinodynamic trajectories under robot dynamics, we instead focus on incorporating kinodynamic constraints and physical stability into the TAMP problem. 



\section{PROBLEM FORMULATION} \label{sec:problem}

We formulate our fully observable task and motion planning problem by extending the formulation in \cite{kwon2025fast}:
\begin{equation}\label{eq:tampform}
P \equiv \langle \mathcal{H}, \mathcal{O}, \mathcal{A}, \mathcal{T}, h_0, \mathcal{G} \rangle.
\end{equation}
Here, $\mathcal{H} = \mathcal{S} \times \mathcal{X}$ denotes the {\it hybrid state space}, and $\mathcal{O}$ is a finite set of environment objects.
$\mathcal{S}$ is a finite set of symbolic states, where each symbolic state $s \in \mathcal{S}$ consists of PDDL \cite{fox2003pddl2} predicates such as \texttt{(on bacon stove)}.
$\mathcal{X}$ denotes the set of continuous world states (\eg object poses, robot joint states, etc.).
If $h=(s,x) \in \mathcal{H}$, then $x$ must be \emph{consistent} with $s$, meaning that the continuous state $x$ satisfy the symbolic predicates in $s$.
For instance, if $s = \{\texttt{(clear red)},\allowbreak \texttt{(on red green)},\allowbreak \texttt{(on-table green)},\allowbreak \texttt{(arm-empty)}\}$,
then in $x$, the red block must not have any object on top, its position must be located on top of the green block, the green block must be positioned on the table in a physically valid configuration, and no object should have contact with the arm gripper.

$\mathcal{A}$ denotes the set of possible actions. An action $a(\delta, \kappa) \in \mathcal{A}$ is defined by an action name $a$ (\eg \texttt{pickup}, \texttt{putdown} in the PDDL domain), discrete parameters $\delta$ referring to objects and regions, and continuous parameters $\kappa$ like goal pose, robot configurations and motion trajectory.
$\mathcal{T}: \mathcal{H} \times \mathcal{A} \rightarrow \mathcal{H}$ is the state transition function, 
and $h_0 = (s_0, x_0)\in \mathcal{H}$ is the initial state. 
The set of goal states is defined as
\begin{equation}
\mathcal{G} = \{(s, x) \in \mathcal{H} \mid s \in S^\star, \, x \in X(s)\},
\end{equation}
where $S^\star \subset \mathcal{S}$ is the set of symbolic goal states, and $X(s) \subset \mathcal{X}$ denotes the set of continuous states consistent with $s$ that satisfy the symbolic goal conditions. 

We assume the availability of samplers, \ie motion planners, that propose candidate parameters $\kappa$, and a physics simulator serving as the transition model $\mathcal{T}$.
Our planning objective is to find a policy 
$\pi = \{a_1, \cdots, a_n \mid \forall a_i \in \mathcal{A}\}$
for $P$ in Eq.~\ref{eq:tampform}, such that successive transition from $h_0$ reaches a goal state $\exists h_n \in \mathcal{G}$ {within a finite time of $n$.}


\section{METHODS} \label{sec:method}
An overview of our method is illustrated in Fig.~\ref{fig:method}.

\subsection{Skeleton Space Generation}
\label{sec:skeleton}

Sequencing-first methods compute a complete symbolic task plan as a skeleton, then attempt to refine it by solving a motion-level constraint satisfaction problem, and find a new skeleton if unsolvable. The chosen plan can be geometrically or physically infeasible—for example, an obstacle may block the path to the target object, or the target object may be in a cluttered environment, making it difficult to grasp.
In order to recover from such dead ends, a TAMP planner must provide additional skeletons as alternatives.
To address this, we adopt a top-$k$ symbolic planner \cite{katz2018novel} to generate diverse symbolic task plans and organize them into a \textit{discrete state graph} $G$ (the blue block in Fig.~\ref{fig:method}). This graph provides a structured representation of the sampled skeleton space, enabling us to explore alternative task-level decisions without restarting the symbolic planner whenever motion refinement fails.

The top-$k$ symbolic planner implements the $K^\ast$ algorithm \cite{aljazzar2011k} on the Fast-Downward planner \cite{helmert2006fast}, a PDDL-based symbolic planner. The objective of top-$k$ symbolic planning is, given a domain PDDL and a problem PDDL, to compute $\Pi = \{\pi_1, \pi_2, \dots, \pi_k\}$, a set of $k$ distinct plans of lowest cost that achieve the PDDL goal set $S^\star$ from the initial state $s_0$. 

Collectively, the set $\Pi$ induces a directed graph $G = (V, E)$, which we call a \textit{discrete state graph}:
\begin{itemize}
    \item Each node $s \in V$ corresponds to a symbolic state represented as a set of PDDL predicates.
    \item Each edge $(s \xrightarrow{\,a} s') \in E$ corresponds to a grounded symbolic action $a$ that transitions state $s$ to state $s'$.
\end{itemize}

This discrete graph is not only a compact representation of the symbolic search space but also serves as a guide for hybrid state tree expansion. 
As explained in the next section, at each hybrid node, the outgoing edges of the corresponding discrete node define the set of admissible symbolic actions, which are then concretized with continuous parameters.

\subsection{Hybrid State Tree Expansion} \label{hybrid}
In the previous section, we constructed a discrete state graph $G$ using top-$k$ symbolic planning. This directed graph represents discrete states as nodes and discrete actions as edges, where a single state can have multiple parents.
In contrast, two hybrid state nodes may share the same symbolic state but are extremely unlikely to have identical continuous states (\eg object poses, robot configurations), since the continuous space is uncountable. Consequently, when planning in the hybrid state space, it is more natural to represent the search structure as a tree rather than a graph.

The root node of the hybrid state tree $T$, $h_0 = (s_0, x_0)$, is defined by the initial PDDL state $s_0$ specified in the problem PDDL and an initial continuous state $x_0$. $x_0$ is instantiated by randomly sampling object states that satisfy the initial symbolic predicates $s_0$. 
Let the current hybrid node be $h_t = (s_t, x_t)$. Expanding from this node proceeds in three stages (the green block in Fig.~\ref{fig:method}):

\subsubsection{Candidate Action Generation} \label{hybrid1}
We first locate the corresponding discrete node $s_t$ in the discrete state graph $G$ and retrieve its outgoing edges, which represent an applicable set of symbolic actions 
$\{a^1_t(\delta^1), \dots, a^m_t(\delta^m)\}$ for $s_t$. These discrete actions are then refined into instantiated hybrid actions by sampling continuous parameters: $\{a^1_t(\delta^1, \kappa^1), \dots, a^m_t(\delta^m, \kappa^m) \mid \kappa^i = (p^i, q^i, q'^i, \xi^i)\}$
where $p^i \in \mathrm{SE}(3)$ is an end-effector's pose in the workspace (\eg grasp or placement pose) corresponding to the action $a$, $q^i \in \mathcal{C}$ is the current robot configuration in the configuration space $\mathcal{C}$ and $q'^i \in \mathcal{C}$ is the robot configuration for $p^i$, and $\xi^i : [0,1] \rightarrow \mathcal{C}$ is a motion trajectory from $q^i$ to $q'^i$. The refinement process consists of:
\begin{enumerate}
  \renewcommand{\theenumi}{\alph{enumi}}
  \renewcommand{\labelenumi}{(\theenumi)}
    \item \textbf{Goal sampling:} For \texttt{pickup} actions, a grasp pose is sampled based on the object's bounding box, with a randomly sampled roll angle applied to the end-effector frame. For \texttt{putdown} actions, a feasible placement position within the target region and the roll angle of the end-effector frame are sampled randomly. Then collision detection is handled by an analytic narrow-phase solver for box–box collisions after an AABB-based sweep-and-prune broad phase by \cite{Genesis}.
    \item \textbf{Inverse kinematics (IK):} For each sampled goal pose $p^i$, an IK solver computes a feasible robot configuration $q'^i$. 
    \item \textbf{Motion planning:} Given the goal configuration $q'^i$, a motion planner computes a collision-free trajectory from $q^i$ to $q'^i$. We use the RRT-Connect planner \cite{kuffner2000rrt}.
\end{enumerate}

If the IK solver fails to return joint positions yielding the desired end-effector pose, or if the motion planner fails to compute a collision-free trajectory, we regard this as a violation of \textit{kinematic or collision constraints}. When such a violation occurs during the refinement of an action candidate $a^i_t(\delta^i, \kappa^i)$, the action is considered \textit{infeasible}. If all candidate refinements fail, replanning is triggered.

\subsubsection{Candidate Action Simulation} \label{hybrid2}
Each \textit{feasible} candidate action $a^i_t(\delta^i, \kappa^i)$ is executed in the physics simulator, producing a successor hybrid state $h^i_{t+1} = \mathcal{T}(h_t, a^i_t(\delta^i, \kappa^i))$. During execution, the simulator records four exocentric views (front, top, left, right) of the scene for each resulting state. 
We employ \cite{Genesis} as the physics simulator, providing efficient physics simulation and photo-realistic rendering.

To verify inertial and contact dynamics effects from the robot motion, we check whether the resulting continuous state $x^i_{t+1}$ obtained from the simulator is consistent with the symbolic state $s^i_{t+1}$ (\ie$x^i_{t+1} \in X(s^i_{t+1})$). This ensures that the action outcome represents the intended symbolic effect without failures such as grasp slippage, collisions, or object collapse. For example, \textit{grasp constraints} are violated if the object falls out of the gripper, while \textit{release constraints} are violated if the object shows significant displacement after being released. If all candidates fail these checks, replanning is triggered.

\subsubsection{VLM-Guided Selection} \label{hybrid3}
The set of valid successor states is
$
\{\, h^i_{t+1} \equiv (s^i_{t+1}, x^i_{t+1}) = \mathcal{T}(h_t, a^i_t(\delta^i, \kappa^i)), \; \forall i \mid a^i_t \text{ is feasible},\; x^i_{t+1} \in X(s^i_{t+1})\},
$
where only actions that are both feasible and execution-consistent are included.
This set is evaluated using a VLM, which leverages commonsense knowledge to guide the search. Specifically, we prompt the VLM with the simulator-rendered images of the current node $h_t$ and candidate successor states, as well as the problem/domain descriptions. The VLM then selects the most promising successor state to continue the search toward the goal.

This hybrid expansion procedure ensures that symbolic decisions are immediately grounded in geometry and validated in simulation, while the VLM biases exploration toward states that are geometrically and kinodynamically consistent.

\subsection{Replanning}
We invoke a replanning mechanism when all action candidates are infeasible at $h_t$, or when every generated successor state is rejected due to kinodynamic constraint violations. In such cases, the planner attempts to recover through a two-stage strategy.

\subsubsection{Retry with Randomized Sampling}
First, we repeat the refinement process at $h_t$ up to $K$ times (the upper red block in Fig.~\ref{fig:method}). Both goal sampling and motion planning are randomized in candidate action generation, so repeated trials increase the likelihood of success. Although a single invocation of RRT-Connect is not guaranteed to succeed, repeated iterations ensure probabilistic completeness. We fix $K=5$ in our implementation.

\subsubsection{VLM-Guided Backtracking}
If all $K$ retries fail, we invoke a VLM-guided backtracking strategy (the lower red block in Fig.~\ref{fig:method}).
The VLM receives the simulator-rendered images of the current node $h_t$, the goal state, a JSON-encoded representation of the hybrid state tree expanded thus far, and a \textit{constraint violation feedback} of previous expansion attempts.
The JSON tree encodes each node with its name, discrete and continuous states, and each edge with information such as action labels, feasibility, and sampled parameters.

The constraint violation feedback is constructed from the outcomes of goal sampling, IK solver, motion planning, and candidate action simulation. 
Inspired by \cite{wang2024llmˆ} and \cite{curtis2024trust}, we classify the feedback into four categories: (i) when no feasible IK solution exists, (ii) when the goal configuration collides with other objects, (iii) when the motion planner fails to produce a collision-free trajectory, and (iv) when a trajectory is found but grasp or release constraints are violated during execution.
This structured feedback, together with the images of the current node $h_t$, helps the VLM identify the cause of failure using both textual feedback and visual cues. 
Based on this information, the VLM then selects an appropriate node to backtrack to within the JSON tree, say $h_r = (s_r, x_r)$.
From $h_r$, step~\ref{hybrid1}, step~\ref{hybrid2}, and step~\ref{hybrid3} in Sec.~\ref{hybrid} are repeated until either a goal state $h_g \in \mathcal{G}$ is reached or the timeout is exceeded.

By combining randomized retries with VLM-guided backtracking, our framework efficiently recovers from unexpandable nodes and explores promising alternative states in the large hybrid search space. For example, suppose repeated attempts in the same region are likely to fail, such as in a cluttered workspace or blocked access. In that case, the VLM can backtrack to a node that enables clearing or repositioning other objects first.

\section{EXPERIMENTS}  \label{sec:experiments}

\begin{table*}[t]
\centering
\resizebox{\textwidth}{!}{%
\begin{tabular}{@{}cccccccccc@{}}
\toprule
\multicolumn{2}{c}{{\color[HTML]{000000} Domain}}                                                                                                     & \multicolumn{4}{c}{{\color[HTML]{000000} \textbf{Blocksworld}}}                                                                                                                               & \multicolumn{4}{c}{{\color[HTML]{000000} \textbf{Kitchen}}}                                                                                                             \\ \midrule
\multicolumn{2}{c|}{{\color[HTML]{000000} $n$}}                                                                                                         & {\color[HTML]{000000} 3}                & {\color[HTML]{000000} 4}                 & {\color[HTML]{000000} 5}                 & \multicolumn{1}{c|}{{\color[HTML]{000000} 6}}                 & {\color[HTML]{000000} 3}                & {\color[HTML]{000000} 4}                & {\color[HTML]{000000} 5}                 & {\color[HTML]{000000} 6}                 \\ \midrule
\multicolumn{1}{c|}{{\color[HTML]{000000} }}                                          & \multicolumn{1}{c|}{{\color[HTML]{000000} SR (\%)}}           & {\color[HTML]{000000} \textbf{100}}              & {\color[HTML]{000000} \textbf{90}}               & {\color[HTML]{000000} \textbf{100}}                & \multicolumn{1}{c|}{{\color[HTML]{000000} \textbf{80}}}                & {\color[HTML]{000000} \textbf{90}}               & {\color[HTML]{000000} \textbf{100}}              & {\color[HTML]{000000} \textbf{90}}                & {\color[HTML]{000000} \textbf{100}}               \\
\multicolumn{1}{c|}{\multirow{-2}{*}{{\color[HTML]{000000} Ours}}}                    & \multicolumn{1}{c|}{{\color[HTML]{000000} Planning Time (s)}} & {\color[HTML]{000000} 67.4221} & {\color[HTML]{000000} 98.1700} & {\color[HTML]{000000} 189.3949} & \multicolumn{1}{c|}{{\color[HTML]{000000} 267.3389}} & {\color[HTML]{000000} 85.6074} & {\color[HTML]{000000} \textbf{93.5570}} & {\color[HTML]{000000} 108.3262} & {\color[HTML]{000000} 120.2675} 
\\ \midrule
\multicolumn{1}{c|}{{\color[HTML]{000000} }}                                          & \multicolumn{1}{c|}{{\color[HTML]{000000} SR (\%)}}           & {\color[HTML]{000000} 100}               & {\color[HTML]{000000} 80}                & {\color[HTML]{000000} 70}                & \multicolumn{1}{c|}{{\color[HTML]{000000} 50}}                & {\color[HTML]{000000} 90}               & {\color[HTML]{000000} 80}               & {\color[HTML]{000000} 80}                & {\color[HTML]{000000} 100}               \\
\multicolumn{1}{c|}{\multirow{-2}{*}{{\color[HTML]{000000} Ours w/o VLM Backtrack}}} & \multicolumn{1}{c|}{{\color[HTML]{000000} Planning Time (s)}} & {\color[HTML]{000000} 55.4456}          & {\color[HTML]{000000} 108.0900}          & {\color[HTML]{000000} 170.0252}          & \multicolumn{1}{c|}{{\color[HTML]{000000} \textbf{254.2184}}}          & {\color[HTML]{000000} 80.4048}          & {\color[HTML]{000000} 95.8650}          & {\color[HTML]{000000} 110.1202}          & {\color[HTML]{000000} \textbf{119.0159}}          \\ \midrule
\multicolumn{1}{c|}{{\color[HTML]{000000} }}                                          & \multicolumn{1}{c|}{{\color[HTML]{000000} SR (\%)}}           & {\color[HTML]{000000} 90}               & {\color[HTML]{000000} 60}                & {\color[HTML]{000000} 30}                & \multicolumn{1}{c|}{{\color[HTML]{000000} 0}}                & {\color[HTML]{000000} 30}               & {\color[HTML]{000000} 0}               & {\color[HTML]{000000} 0}                 & {\color[HTML]{000000} 0}                 \\
\multicolumn{1}{c|}{\multirow{-2}{*}{{\color[HTML]{000000} PDDLStream}}}              & \multicolumn{1}{c|}{{\color[HTML]{000000} Planning Time (s)}} & {\color[HTML]{000000} \textbf{0.8247}}           & {\color[HTML]{000000} \textbf{14.0807}}           & {\color[HTML]{000000} 162.3950}          & \multicolumn{1}{c|}{{\color[HTML]{000000} Timeout}}          & {\color[HTML]{000000} \textbf{43.6913}}          & {\color[HTML]{000000}Timeout}          & {\color[HTML]{000000} Timeout}           & {\color[HTML]{000000} Timeout}           \\ \midrule
\multicolumn{1}{c|}{{\color[HTML]{000000} }}                                          & \multicolumn{1}{c|}{{\color[HTML]{000000} SR (\%)}}           & {\color[HTML]{000000} 100}              & {\color[HTML]{000000} 80}                & {\color[HTML]{000000} 70}                & \multicolumn{1}{c|}{{\color[HTML]{000000} 30}}                & {\color[HTML]{000000} 90}               & {\color[HTML]{000000} 0}                & {\color[HTML]{000000} 10}                & {\color[HTML]{000000} 0}                 \\
\multicolumn{1}{c|}{\multirow{-2}{*}{{\color[HTML]{000000} LLM$^3$}}}                    & \multicolumn{1}{c|}{{\color[HTML]{000000} Planning Time (s)}} & {\color[HTML]{000000} 6.4879}           & {\color[HTML]{000000} 74.0670}           & {\color[HTML]{000000} \textbf{132.2868}}          & \multicolumn{1}{c|}{{\color[HTML]{000000} 270.5547}}          & {\color[HTML]{000000} 84.9920}          & {\color[HTML]{000000} Timeout}          & {\color[HTML]{000000} \textbf{67.7273}}           & {\color[HTML]{000000} Timeout}           \\ \bottomrule
\end{tabular}%
}
\caption{Average success rates (\%) and planning times (s) of all baselines for $3 \leq n \leq 6$. Planning times are averaged over successful trials only. For cases with 0\% success, the planning time is reported as "Timeout".}
\label{tab:table1}
\vspace{-1.5em}
\end{table*}

\subsection{Experimental Setup}

All TAMP experiments were conducted on a workstation with a 13th Gen Intel Core i9-13900K CPU and an NVIDIA GeForce RTX 4090 GPU. 
We employed GPT-4o \footnote{\url{https://openai.com}} from OpenAI as the VLM, with temperature fixed to 0.0. 
For generating symbolic plan skeletons, we used a top-$k$ symbolic planner \cite{katz2018novel}, with the number of plans fixed to $k=30$.  
Genesis \cite{Genesis} was used as the physics simulator and photo-realistic renderer.
A planning timeout of 600 seconds was imposed for all experiments, with planning proceeding until a goal was found or the timeout was reached. After the planning was terminated, the resulting simulator state was checked against the PDDL goal conditions; if satisfied, the trial was marked as a success; otherwise, a failure.

We evaluated our approach in two simulated domains with varying numbers of objects $n$, which determine problem complexity:

\begin{figure}
    \centering
    \includegraphics[width=\columnwidth]{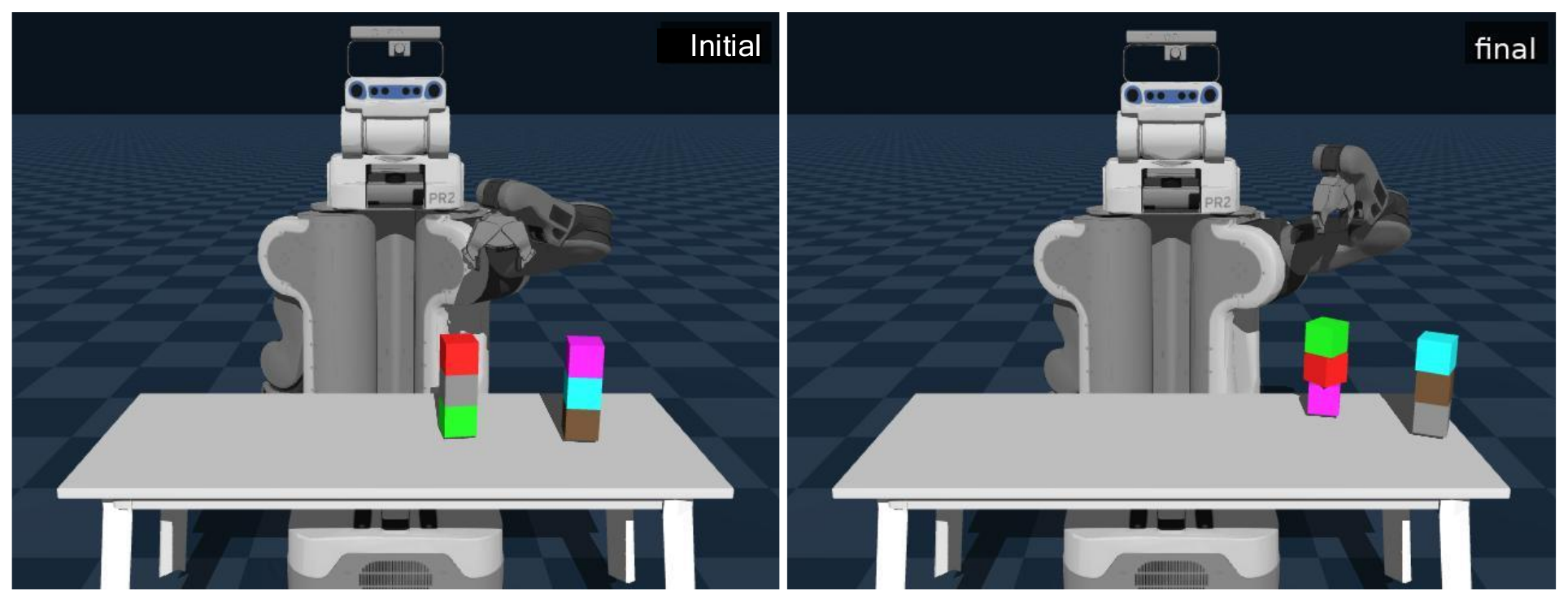}
    \caption{Blocksworld domain setup with a PR2 robot. We only enable the left arm for this task. Initially, $n=6$ blocks are randomly placed on a table (left), and the goal is to restack them in a new order (right).}
    \label{fig:blocksworld}
    \vspace{-1.0em}
\end{figure}

\begin{figure}
    \centering
    \includegraphics[width=\columnwidth]{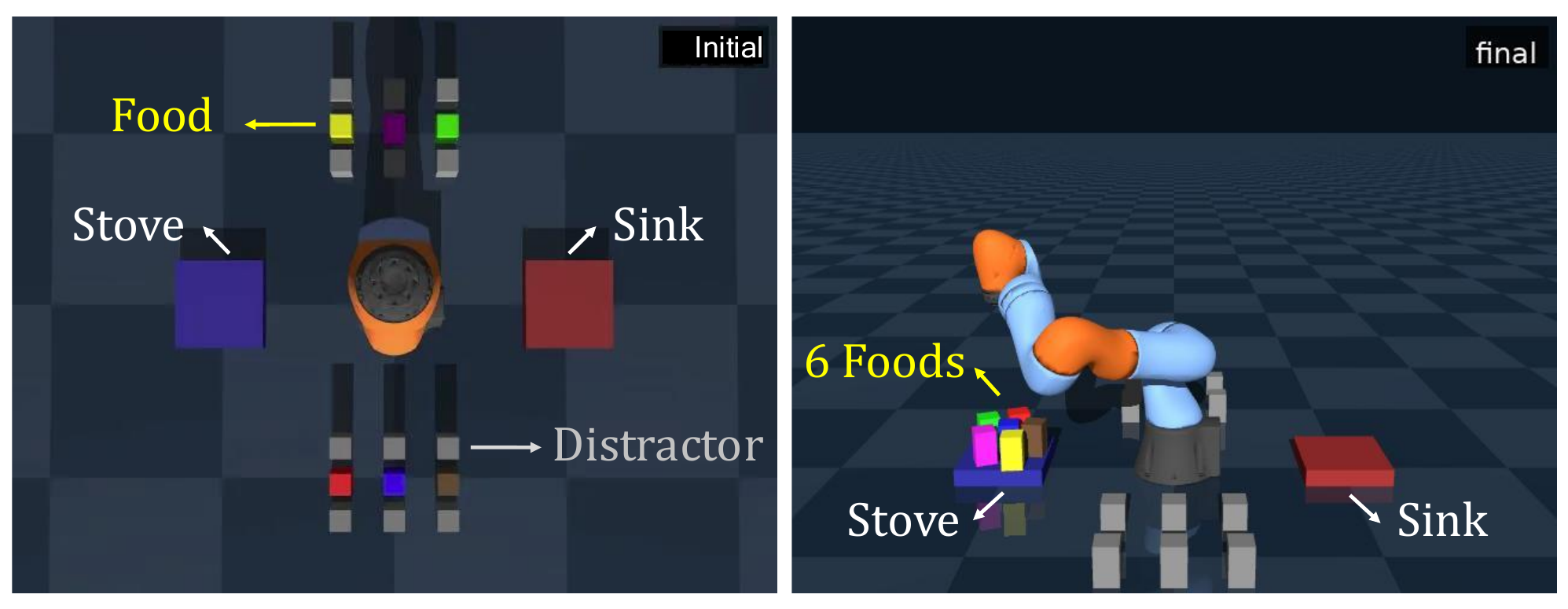}
    \caption{Kitchen domain setup with a KUKA IIWA robot. Initially, six food objects are surrounded by 12 distractors (left), and the goal is to cook the food on the stove (right). The distractors are fixed objects.}
    \label{fig:kitchen}
    \vspace{-1.5em}
\end{figure}

\begin{itemize}
    \item \textbf{Blocksworld:} This domain is based on the PDDL benchmark domain \cite{seipp-et-al-zenodo2022} from the International Planning Competitions (IPC). As in Fig.~\ref{fig:blocksworld}, $n$ blocks are randomly divided into 2-3 stacks on the table, and the goal is to restack them in a new target order. 
    Available actions include \texttt{pickup}, \texttt{putdown}, \texttt{stack}, and \texttt{unstack}.
    
    \item \textbf{Kitchen:} This domain is a modified version of the Kitchen domain used in PDDLStream \cite{garrett2020pddlstream}. As in Fig.~\ref{fig:kitchen}, it contains 6 colored food objects (radish, egg, bacon, chicken, celery, apple), a sink (red cuboid), and a stove (blue cuboid). A food block becomes \texttt{Cooked} only after being cleaned on the sink and then cooked on the stove. 
    The goal is to make a randomly chosen $n$ food objects into \texttt{Cooked} state. To increase problem difficulty, we added gray distractor blocks around the initial positions of the food objects. 
\end{itemize}

\subsection{Performance Analysis}
For both domains, we varied $3\le n \le 6$. For each value of $n$, we randomly generated 10 problem instances and measured both success rate and planning time.
Table.~\ref{tab:table1} shows the results for each domain. 
We quantitatively compared 4 approaches:

\begin{enumerate}
    \item \textbf{Ours}: Our proposed method with VLM-guided backtracking, as introduced in Sec.~\ref{hybrid}.
    \item \textbf{Ours w/o VLM Backtrack}: Our method without VLM guidance for backtracking. When a failure occurs, we return to a simple BFS-based strategy that selects the first unvisited discrete node and resumes search from the corresponding hybrid node.
    \item \textbf{PDDLStream}\cite{garrett2020pddlstream}: A domain-independent traditional TAMP baseline that integrates symbolic planning with black-box sampling functions {(\eg grasp sampler, IK solver, RRT-Connect motion planner, etc.)}. We compare against its "adaptive" algorithm. Search-sample ratio is fixed to 1.
    \item \textbf{LLM$^3$}\cite{wang2024llmˆ}: An LLM-based TAMP baseline where the LLM directly predicts symbolic action sequences and continuous parameters. We compare against the variant "LLM$^3$ Backtrack", where the LLM selects a previous action for backtracking and replans from that point upon failure.
\end{enumerate}

{\bf Comparisons:} 
Our method outperforms baselines, achieving an average success rate of 92.5\% in Blocksworld and 95\% in Kitchen. In contrast, PDDLStream reports success rates of only 45\% and 7.5\% in Blocksworld and Kitchen, respectively. Moreover, its planning time grows exponentially as $n$ increases, eventually leading to timeouts at $n=6$ in Blocksworld, $n=4,5,6$ in Kitchen for all ten problem instances.
PDDLStream employs streams, which are black-box sampling functions that generate continuous parameters and certify constraints, such as collision-free motion. 
The outputs of streams, called "stream objects", are added to the symbolic state and reused when searching for new plans. 
As more stream objects accumulate, the planning state space grows significantly, slowing the process and often leading to inefficiency \cite{vu2024coast}. 

LLM$^3$ achieved average success rates of 70\% in Blocksworld and 25\% in Kitchen, but suffered timeouts at $n=4,6$ in Kitchen for all problem instances and at $n=5$ for 90\% of problem instances. 
In Blocksworld, the success rate drops at $n=6$, where the LLM increasingly predicts misordered stacking sequences, resulting in an average of 42 backtrack trials. In Kitchen, failures arise from collisions when placing food blocks on the stove, as LLMs struggle to predict collision-free poses and often continue producing infeasible continuous action parameters outputs even after replanning.

The comparison with the baselines clarifies the role of each component in our pipeline. 
The discrete state graph generated by the top-$k$ planner explicitly bounds the symbolic search space, preventing the planning-time explosion caused by uncontrolled search space growth observed in PDDLStream. 
Moreover, hybrid state tree expansion with motion planning and physics-based rollout grounds each action in geometric and dynamic feasibility, mitigating the geometric reasoning limitations of purely LLM-based TAMP planners.

{\bf Ablation Study of VLM-guided Backtracking:} 
We conducted an ablation study to evaluate the effectiveness of VLM-guided backtracking. As shown in Table~\ref{tab:table1}, our method with VLM-guided backtracking achieved 23.33\% and 8.57\% increases in average success rates in Blocksworld and Kitchen, respectively, compared to the version without VLM guidance. 
This is because VLMs leverage commonsense knowledge and visual reasoning to identify the cause of failure in the current state and backtrack to a feasible node within one or two attempts. In contrast, without VLM-guided backtracking, the planner often repeatedly attempted backtracking until reaching a timeout.
While incorporating VLM-guided backtracking slightly increased planning time due to additional VLM calls, the overall difference was marginal.

{\bf Large Task Space vs. Large Motion Space:} 
Blocksworld exemplifies a \textit{large task space}, as its complexity arises from the high branching factor induced by arbitrary block stacking configurations and discrete actions such as \texttt{stack} and \texttt{unstack} \cite{khodeir2023learning, lagriffoul2018platform}.
By contrast, the Kitchen domain is characterized by a \textit{large motion space}, where tight manipulation constraints make it challenging to grasp food objects in cluttered environments with distractors and to place multiple items on the stove or sink without collisions \cite{ren2024extended, lagriffoul2018platform}.

Our ablation also confirms this distinction: VLM-guided backtracking yields a larger performance gain in Blocksworld than in Kitchen. 
In Blocksworld, where difficulty is primarily due to symbolic branching, VLM backtracking effectively explores alternative discrete states and mitigates task-level search complexity.
In the Kitchen, however, the primary challenge lies in continuous motion feasibility under cluttered initial positions and constrained placements, so backtracking to a different discrete action has less impact.
This demonstrates that VLM-guided backtracking is particularly effective in domains characterized by large task spaces, where symbolic branching dominates overall complexity.


\begin{figure*}
    \centering
    \includegraphics[width=\textwidth]{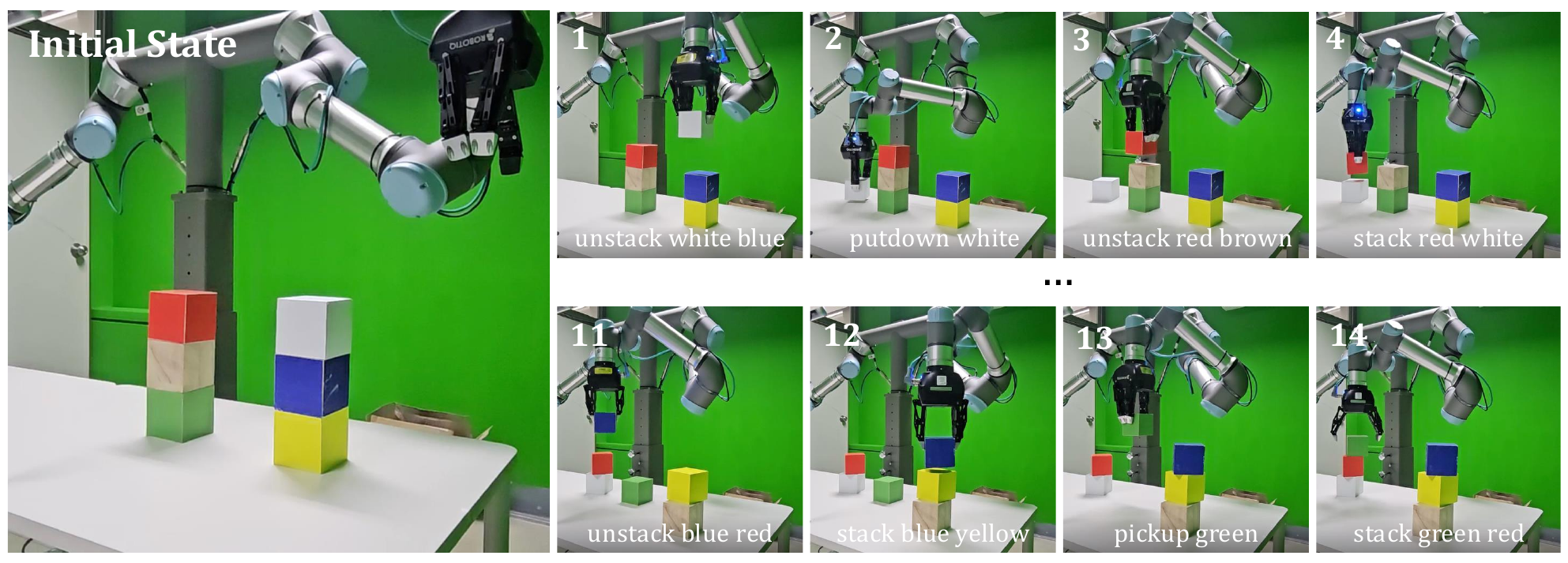}
    \caption{Robotic demonstration of our TAMP planner in the Blocksworld domain ($n=6$). The initial configuration consists of six blocks stacked on the table (leftmost image), and the goal is to rearrange them into the stacking sequence shown in 14. Only the first four and last four actions, $\pi = \{a_1, a_2, a_3, a_4, \cdots, a_{11}, a_{12}, a_{13}, a_{14}\}$, are shown here.}
     \label{fig:real}
    \vspace{-1.0em}
\end{figure*}

\subsection{Real-World Demonstration}
We further validated our approach in Blocksworld through a real-world robotic demonstration. The setup consisted of dual UR5e manipulators equipped with Robotiq 3F grippers and an Intel RealSense D455 sensor. Given an RGB image of the tabletop scene captured by the sensor, an open-vocabulary object detection model \cite{ren2024grounded} generated segmented masks for each object. These 2D masks were projected into 3D points using depth values, from which each object was localized in the simulator. 
Since the geometric and inertial properties of the objects were assumed to be known a priori, planning was performed as described in Sec.~\ref {sec:method}.
The resulting plan, consisting of the arm trajectories and gripper open/close actions, was directly executed on the real robot, as shown in Fig.~\ref{fig:real}.

We conducted 5 trials for each $3\le n \le 6$ in the Blocksworld domain. 
The planning times and success rates are similar to the simulation results presented in Sec.~\ref {sec:experiments}. In particular,
\begin{itemize}
\item For $n=3$ and $n=4$, the success rates are $100 \%$.
\item For $n=5$, two trials failed due to collisions between the gripper and the object, due to the inaccurate object localization caused by occlusion.
\item For $n=6$, the success rates are $80 \%$, similar to the simulation.
\end{itemize}
These results demonstrate that our planning approach can generate feasible plans for long-horizon tasks applicable in real-world settings.


\section{CONCLUSION AND FUTURE WORK} \label{sec:conclusion}
This paper presents a novel kinodynamic TAMP planner that interleaves symbolic planning with kinodynamic feasibility checks using a physics simulator as the world transition model, while a VLM guides forward search and replanning.

{\bf Limitations:}
A major limitation of this work is that the computational overhead introduced by physics simulation increases with plan length.
Another limitation is that performance can be sensitive to the quality of the underlying samplers and variations in the VLM, such as different models, temperature settings, prompt design, and the number of input images. 
In addition, the assumptions of full observability and a black-box transition model remain restrictive.

{\bf Future Work:}
Future work will focus on extending our approach beyond pick-and-place-style manipulation to more diverse domains such as tool use, deformable objects, and contact-rich tasks. These domains require more expressive abstractions and validation strategies, and could be evaluated on more realistic benchmarks such as \cite{liu2023libero}. 
We can also reduce sensitivity to sampler quality by integrating learned sampling strategies to improve efficiency, as well as to relax assumptions such as full observability.

\section*{ACKNOWLEDGMENT}
This work was supported in part by the ITRC/IITP Program (IITP-2026-RS-2020-II201460), and in part by the NRF (NRF-2022R1A2B5B03001385) in South Korea.









\bibliographystyle{IEEEtran}
\bibliography{IEEEabrv,main}

\end{document}